\documentclass[runningheads]{llncs}
\usepackage{graphicx}
\usepackage{tikz}
\usepackage{comment}
\usepackage{amsmath,amssymb} % define this before the line numbering.
\usepackage{color}
\usepackage{threeparttable}
\usepackage{multirow}
\usepackage{array}
\usepackage{subfig}
\usepackage{graphicx} 
\usepackage{wrapfig}
\usepackage{pifont}
\usepackage{booktabs}       % professional-quality tables
\usepackage[breaklinks=true,bookmarks=false]{hyperref}
\hypersetup{
    colorlinks=true,
    citecolor=darkblue
}
\definecolor{darkblue}{rgb}{0,0.08,0.45}

\usepackage{bbding}
\usepackage{graphicx, amsmath, amssymb, caption, multirow, overpic, textpos}
\usepackage{pifont}
\usepackage{adjustbox}
\newcommand{\cmark}{\ding{51}}
\newcommand{\xmark}{\ding{55}}
\usepackage[accsupp]{axessibility}  % Improves PDF readability for those with disabilities.

\makeatletter
\def\thickhline{%
	\noalign{\ifnum0=`}\fi\hrule \@height \thickarrayrulewidth \futurelet
	\reserved@a\@xthickhline}
\def\@xthickhline{\ifx\reserved@a\thickhline
	\vskip\doublerulesep
	\vskip-\thickarrayrulewidth
	\fi
	\ifnum0=`{\fi}}
\makeatother

\newlength{\thickarrayrulewidth}
\begin{document}
% \pagestyle{heading}
% \mainmatter

\title{Beyond Masking: Demystifying Token-Based Pre-Training for Vision Transformers}

\titlerunning{Beyond Masking: Demystifying Token-Based Pre-Training for Vision Transformers}

\author{Yunjie Tian\inst{1} \and
Lingxi Xie\inst{2} \and
Jiemin Fang\inst{3} \and
Mengnan Shi\inst{1} \and
Junran Peng\inst{4} \and
Xiaopeng Zhang\inst{2} \and
Jianbin Jiao\inst{1} \and
Qi Tian\inst{2} \and
Qixiang Ye\inst{1}}
\authorrunning{Y. Tian et al.}
\institute{University of Chinese Academy of Sciences \and Huawei Inc. \and
Institute of AI \& School of EIC, Huazhong University of Science and Technology \and
Chinese Academy of Sciences}

%******************
\maketitle

\begin{abstract}
The past year has witnessed a rapid development of masked image modeling (MIM). MIM is mostly built upon the vision transformers, which suggests that self-supervised visual representations can be done by masking input image parts while requiring the target model to recover the missing contents. MIM has demonstrated promising results on downstream tasks, yet we are interested in whether there exist other effective ways to `learn by recovering missing contents'. In this paper, we investigate this topic by designing five other learning objectives that follow the same procedure as MIM but degrade the input image in different ways. With extensive experiments, we manage to summarize a few design principles for token-based pre-training of vision transformers. In particular, the best practice is obtained by keeping the original image style and enriching spatial masking with spatial misalignment -- this design achieves superior performance over MIM in a series of downstream recognition tasks without extra computational cost. The code is available at
\href{https://github.com/sunsmarterjie/beyond\_masking}{\color{magenta}https://github.com/sunsmarterjie/beyond\_masking}.

\end{abstract}

\section{Introduction}
\label{introduction}

Transformers~\cite{vit,swin,zhou2021deepvit,wang2021pyramid,chen2021crossvit} have shown their great potentials in visual recognition, but these models often require a large amount of labeled training data. To alleviate the burden, it is natural to consider self-supervised learning, \textit{i.e.}, pre-training vision transformers without semantic labels. Self-supervised learning is a long-lasting battle in the computer vision community, where various approaches have been proposed, and typical learning proxies include generating image contents~\cite{inpainting,sage}, determining image geometries~\cite{jigsaw,context}, pushing the consistency between multiple views~\cite{moco,byol,moco-v2}, \textit{etc}. The last type, often known as contrastive learning, contributed successful algorithms on vision transformers~\cite{moco-v3,dino}.

Vision transformers maintain a set of tokens to extract visual features. Recently, researchers proposed token-based learning objectives that investigated the behavior of each tokens rather than the entire image -- as contrastive learning approaches did. A representative methodology is known as masked image modeling (MIM)~\cite{beit,mae,CAE}, where a part of image patches are removed from input and the goal is to recover the missing contents or some kind of statistics~\cite{maskfeat}. These approaches are closely related to the generation-based proxies, yet they take advantage of the property of vision transformers that tokens are relatively individual from each other (unlike the convolutional networks in which the relationship between neighboring neurons is much closer than that of distant neurons). In practice, MIM achieves promising visual recognition accuracy.

MIM defines a general framework that first degrades the input image (\textit{i.e.}, partial information is missing) and then requires the target model to recover the missing contents -- masking is a particular way of degradation. We are interested in the following questions: \textbf{(1) Is masking the only effective way of degradation? (2) More importantly, are there any design principles for degradation?} To answer the questions, we design five other learning objectives (\textit{i.e.}, pre-training tasks), all of which follow the general framework but the degradation is performed differently -- in particular, the input image is (a) a \texttt{zoomed-in} image, (b) a \texttt{zoomed-out} image, (c) a \texttt{distorted} image, (d) a \texttt{blurred} image, and (e) a \texttt{de-colorized} image. On the standard evaluation protocols (\textit{i.e.}, pre-training the vision transformers on ImageNet and fine-tuning them on ImageNet, COCO, and ADE20K), we compare these five tasks to MIM and the random baseline (\textit{i.e.}, the weights have not been pre-trained), and deliver the following insightful messages.

\textbf{First}, as expected, MIM as well as all five designed tasks surpass the random baseline, implying that token-based image recovery is more or less helpful to visual representation learning. Some of these pre-training tasks achieve comparable performance with MIM. Surprisingly integrating the introduced task with masking ones can further boost the performance.
% Surprisingly, some of these pre-training tasks (\textit{e.g.}, recovering from a \texttt{distorted} image), while having removed much fewer information from the input image, is only slightly inferior to MIM in terms of downstream recognition accuracy.

\textbf{Second}, to further explain the empirical results by decomposing MIM as well as the designed pre-training tasks into three degradation factors: (i) part of input information is missing; (ii) the input image is spatially transformed; (iii) the input image style has been changed. Note that each pre-training task may own multiple factors. By comparison, we find that all three factors are helpful to visual representation learning, but the last factor (\textit{i.e.}, changing the image style) is prone to changing the input data distribution and thus incurs a domain gap between the upstream pre-training and downstream fine-tuning -- this leads to unsatisfying performance of pre-trained models.

\textbf{Third}, going one step further, we show that combining spatial masking and spatial misalignment yields stronger learning objectives that produce better pre-training performance. In particular, when the input is obtained by adding masks on a \texttt{zoomed-in} image, the ViT-base model, pre-trained on ImageNet for $300$ epochs, surpasses the MIM counterpart consistently on ImageNet classification (accuracy), COCO object detection (box AP), and ADE20K semantic segmentation (mIOU), respectively. This indicates that ample room exists for token-based pre-training in the future.

\textbf{Lastly}, in Section~\ref{discussions}, we offer remarks on token-based pre-training, including comparing it to image-based pre-training (mostly contrastive learning) and summarize its limitations. We hope that our study as well as discussions can shed light to future research of pre-training vision transformers.

\section{Related Work}
\label{relatedwork}

In the deep learning~\cite{lecun2015deep} era, deep neural networks play an important role in visual recognition. Originally, convolution is the key module (that contributes most parameters and computations) to build neural architectures~\cite{alexnet,googlenet,vggnet,resnet,densenet}, recently, researchers have transplanted a powerful module from natural language processing named transformers~\cite{all-you-need}. Different from convolution-based networks that scan the feature map using a sliding window, vision transformers~\cite{vit} maintain a set of tokens and compute attentions between the tokens for feature extraction. Follow-up efforts have largely improved vision transformers in terms of data~\cite{deit,zhou2021deepvit} and computational~\cite{mobilevit,graham2021levit,wang2022pvt} efficiency. Some works~\cite{swin,chu2021Twins,fang2021msgtransformer,dong2021cswin,huang2021shuffle,yu2021glance} build vision transformers into a hierarchical framework which significantly boost performance on many vision tasks including image classification, object detection, semantic/instance segmentation, \textit{etc}.

The requirement of abundant training data is a major burden of vision transformers. To alleviate it, a promising path is to perform self-supervised visual representation learning (SSVRL), or briefly referred to as unsupervised pre-training in this paper. The key to SSVRL is the pre-training proxy, \textit{i.e.}, a learning objective (without semantic labels) that forces the deep neural networks to adjust the weights. We categorize the learning objectives into three parts. \textbf{(A)} The early efforts date back to training autoencoders~\cite{hinton2006reducing} that compressed an image into short vectors and required the target model to recover the original contents. Autoencoder has been the representative work of generation-based learning, where later efforts improved the original setting by predicting (partly) missing contents of the input image~\cite{inpainting,yu2018inpainting} or predicting visual primitives~\cite{sage}. \textbf{(B)} The second path emerges when researchers encoded images or patches into geometric codes (\textit{e.g.}, containing spatial contexts~\cite{noroozi2016jgsaw} or spatial transformations~\cite{doersch2015unsupervised,fang2022corrupted}) and required the target model to predict the codes based on the degenerated and/or transformed inputs. These methods were later validated to have limited abilities of learning higher-level visual features~\cite{caron2018deep}. \textbf{(C)} Currently, one of the most effective learning objectives is named contrastive learning~\cite{simclr,moco}, where the goal is to predict whether different views or patches belong to the same image. There are two typical ways to implement contrastive learning: the first one~\cite{simclr,bachman2019learning,van2018representation} built a large memory bank for instance discrimination, and the second one~\cite{byol,moco} trained a mapping function between multiple views to enable them to predict the high-level statistics of each other. On the basis of convolutional neural networks, contrastive learning (without semantic labels) has shown comparable performance with supervised learning (with semantic labels), in terms of visual recognition accuracy in downstream tasks.

It is natural to apply SSVRL approaches to vision transformers, where some successful efforts have been made upon contrastive learning~\cite{moco,moco-v2,simclr}. More interestingly, vision transformers are different from convolution-based networks in terms of how basic units interact with each other. Convolution-based networks use sliding windows to traverse the feature map, so that neighboring neurons are often highly correlated; by contrast, vision transformers (especially the vanilla version~\cite{vit}) only allow tokens to interact with each other under the self-attention mechanism -- in other words, in vision transformers, the basic units are relatively individual compared to that in convolution-based networks. This property gives birth to a new learning objective named masked image modeling (MIM)\footnote{MIM is closely related to masked language modeling (MLM) that were widely used in pre-training NLP models~\cite{bert,xlnet,gpt3}, but we notice that texts are easily decomposed into semantic units (\textit{i.e.}, words), yet images do not enjoy the same benefit (pixels are insufficient to capture semantics).}, in which part of image patches (corresponding to some tokens) are masked and the target model is required to recover the original image patches~\cite{mae} or statistics~\cite{beit,maskfeat} from the incomplete inputs. MIM was verified effective in pre-training vision transformers, especially when the target model is allowed to be fine-tuned in the downstream scenarios. Essentially, MIM belongs to the family of generative learning, but compared to the earlier efforts upon convolution-based networks (in particular, the learning objective by inpainting an image~\cite{inpainting}), the downstream recognition performance is significantly improved. \textbf{The goal of this paper is to demystify where the improvement comes from and reveal the extent of effectiveness of the token-based pre-training approaches.}

\section{Methodology}
\label{methodology}

\subsection{Pipeline: Pre-Training and Fine-Tuning}
\label{methodology:pipeline}

Let $\mathbf{x}$ be an image without semantic labels and $f^\mathrm{B}\!\left(\cdot\right)$ be the target model which maps $\mathbf{x}$ into mid-level features $\mathbf{z}=f^\mathrm{B}\!\left(\mathbf{x};\boldsymbol{\theta}^\mathrm{B}\right)$, where the superscript `$\mathrm{B}$' represents `backbone', and $\boldsymbol{\theta}$ indicates the learnable parameters. To be concrete, $f^\mathrm{B}\!\left(\cdot\right)$ only contains the part that is transferred from upstream (pre-training) to downstream (fine-tuning), \textit{e.g.}, both BEIT~\cite{beit} and MAE~\cite{mae} used the ViT~\cite{vit} architecture, where $f^\mathrm{B}\!\left(\cdot\right)$ contains a \textcolor{black}{projector} and $12$ transformer blocks (which are often referred to as the \textit{encoder}). That said, different `head' networks are needed in the pre-training and fine-tuning procedures, and we denote them as $f^\mathrm{PT}\!\left(\cdot\right)$ and $f^\mathrm{FT}\!\left(\cdot\right)$, where $\mathrm{PT}$ and $\mathrm{FT}$ stand for `pre-training' and `fine-tuning', respectively. For the pre-training task, the goal is to predict $\hat{\mathbf{x}}=f^\mathrm{PT}\!\left(\mathbf{z};\boldsymbol{\theta}^\mathrm{PT}\right)$, some kind of statistics of the original image, while for the fine-tuning task, the goal is to predict $\mathbf{y}=f^\mathrm{FT}\!\left(\mathbf{z};\boldsymbol{\theta}^\mathrm{FT}\right)$, the desired recognition results -- note that we have used different superscripts for both the function and parameters.

Throughout this paper, we consider the context of vision transformers, where $f^\mathrm{B}\!\left(\cdot\right)$ is composed of a set of transformer blocks. The input image, $\mathbf{x}$, is partitioned into $M$ non-overlapped patches, each of which is projected onto a token, and they interact through the self-attention mechanism for $L$ blocks during the forward propagation. We denote $\mathbf{z}_m^{(l)}$ as the $m$-th token ($m=0,1,\ldots,M-1$) after the $l$-th block ($l=0,1,\ldots,L$). As an example, if the ViT-base architecture is used, we have $M=14\times14=196$, $L=12$, $\mathbf{z}_m^{(l)}\in\mathbb{R}^{768}$, and the input image is at $224\times224$ (partitioned into $14\times14$ patches of $16\times16$ RGB values).

The pre-training stage challenges the target model with a learning objective. The objective, sometimes referred to as `the pre-training task', `the proxy task', \textit{etc.}, typically contains two modules for degradation and recovery, respectively. Since semantic information is not provided, the input image is first pre-processed (\textit{e.g.}, partially masked) so that some information is hidden from the input. Then, the target model equipped with the pre-training head, $f^\mathrm{PT}\circ f^\mathrm{B}\!\left(\cdot\right)$, receives the degraded input and tries to recover the missing information. We denote the degraded input as $\tilde{\mathbf{x}}$, which should be strictly distinguished from $\hat{\mathbf{x}}$ that denotes the output, namely, $\hat{\mathbf{x}}=f^\mathrm{PT}\!\left( f^\mathrm{B}\!\left(\tilde{\mathbf{x}};\boldsymbol{\theta}^\mathrm{B}\right);\boldsymbol{\theta}^\mathrm{PT}\right)$. The difference between $\hat{\mathbf{x}}$ and $\mathbf{x}$ is taken as the loss function, \textit{e.g.}, using the mean square error (MSE) criterion, the loss function is written as $\mathcal{L}_\mathrm{MSE}\doteq\left|\hat{\mathbf{x}}-\mathbf{x}\right|_2^2$. After the pre-training stage, the pre-training head $f^\mathrm{PT}\!\left(\cdot;\boldsymbol{\theta}^\mathrm{PT}\right)$ is discarded and the fine-tuning head is appended, yielding in a composed model $f^\mathrm{FT}\circ f^\mathrm{B}\!\left(\cdot\right)$ to be tuned for downstream tasks.

\subsection{MIM and Designed Learning Objectives}
\label{methodology:demystifying}

The key to the above pipeline is the degradation method that produces $\tilde{\mathbf{x}}$ from $\mathbf{x}$. Intuitively, the degradation method determines what and/or how much information has been removed from the input -- in other words, how the target model is challenged to recover the missing contents. The research methodology of this paper involves altering the image degradation method and observe the behavior of the pre-trained models.

A currently popular choice is named masked image modeling (MIM), where the idea is straightforward -- a subset of patches of $\mathbf{x}$ are removed from input, and the model is required to recover the original image from the incomplete information. Mathematically, this is to choose a subset, $\mathcal{M}'\subset\left\{0,1,\ldots,M-1\right\}$, and only the tokens corresponding to the elements in $\mathcal{M}'$ exist in the backbone, \textit{i.e.}, for any $l$, there are $\left|\mathcal{M}'\right|$ tokens in the $l$-th layer. With the mid-level features $\left\{\mathbf{z}_m^{(L)}\right\}$ obtained, a set of dummy tokens (each of them corresponds to a missing element in $\mathcal{M}'$) are appended, followed by the pre-training head to recover the original input. MIM is a special case of using an encoder-decoder architecture for recovering missing image contents, where $f^\mathrm{B}\!\left(\cdot;\boldsymbol{\theta}^\mathrm{B}\right)$ and $f^\mathrm{PT}\!\left(\cdot;\boldsymbol{\theta}^\mathrm{PT}\right)$ serve as the encoder and decoder, respectively. Other designed learning proxies, as described below, follow the same pipeline, despite that some of them do not involve tokens being removed and added back.

\begin{figure}[!t]
\centering
\includegraphics[height=4.8cm]{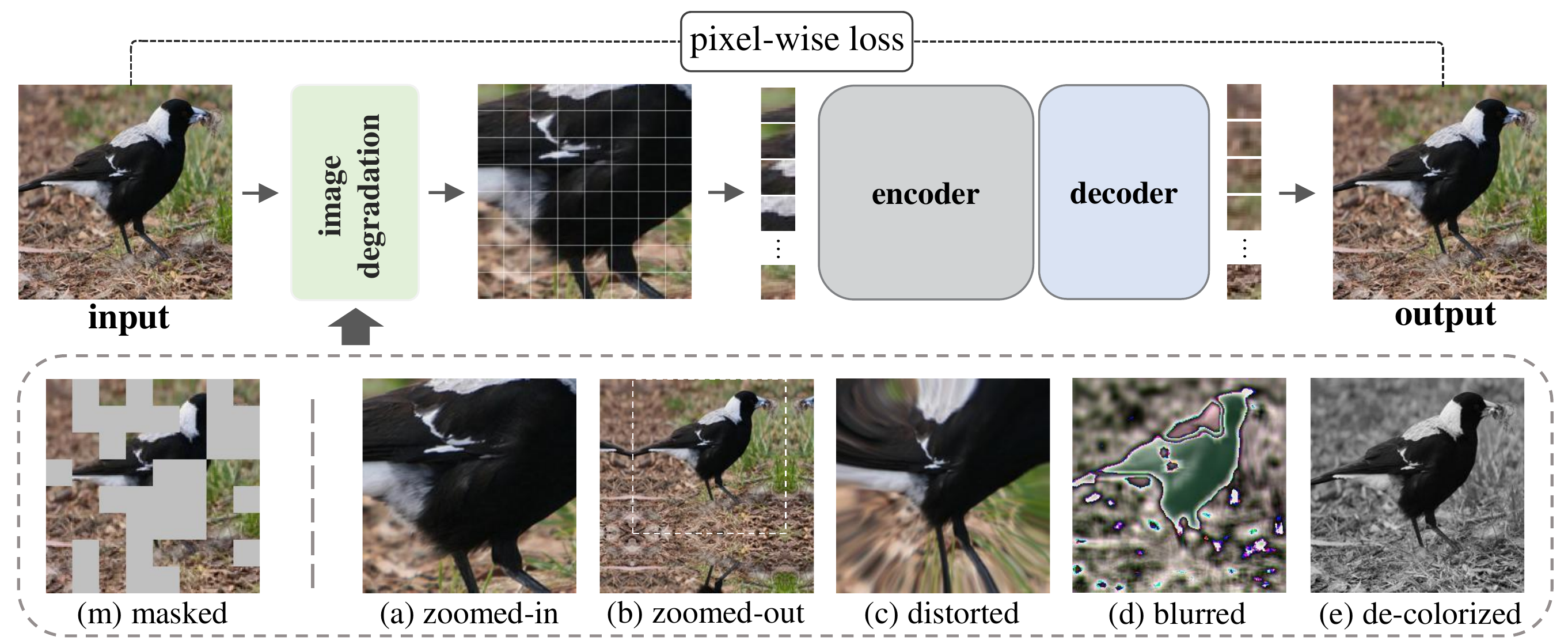}
\caption{Illustration of the research methodology of this paper. There are six candidate learning objectives (please find technical details in the main texts), and each pre-training procedure involves pre-processing the input with one specific degradation and recovering the original image. For better visualization, we partition the degraded image into $8\times8$ patches, yet the actual setting is always $14\times14$. The white square on the \texttt{zoomed-out} input indicate the border of original image. \textit{This figure is best viewed in color.}}
\label{fig:objectives}
\end{figure}

Besides MIM that incurs (i) information missing, a particular type of degradation, we are interested in whether other degradation methods can work. Here, we investigate two common scenarios, namely, (ii) spatial transformation and (iii) image style change, both of which are widely accepted for vision data augmentation. Taking implementation efficiency into consideration, we design five new learning objectives, which require the target model to recover the original image from (a) a \texttt{zoomed-in} image, (b) a \texttt{zoomed-out} image, (c) a \texttt{distorted} image, (d) a \texttt{blurred} image, and (e) a \texttt{de-colorized} image. We show an example of original and degraded images in Figure~\ref{fig:objectives}, and leave the implementation details to the experimental part (see Section~\ref{experiments:pretraining}). Note that each learning objective may correspond to multiple degradation types, \textit{e.g.}, the \texttt{zoomed-in} image is generated by spatial transform, yet part of image contents are missing. Based on these learning objectives, we can observe how different types (i)(ii)(iii) contribute to visual representation learning, and furthermore, construct more effective learning objectives for token-based pre-training of vision transformers.

\section{Empirical Studies}
\label{experiments}

\subsection{Pre-Training Settings and Implementation Details}
\label{experiments:pretraining}

Following the convention, we use the vanilla vision transformer (known as ViT~\cite{vit}) as the backbone network, $f^\mathrm{B}\!\left(\cdot\right)$. We investigate the ViT-base model with $12$ blocks and a channel dimension of $768$. The input, $\mathbf{x}$, is set to be a $224\times224$ image, partitioned into $14\times14$ tokens, each of which sees a $16\times16$ patch on the original image. We pre-train $f^\mathrm{B}\!\left(\cdot\right)$ with the help of $f^\mathrm{PT}\!\left(\cdot\right)$, which has $4$ transformer blocks and the final output has the same dimensionality as the original image. The dataset is ImageNet-1K~\cite{russakovsky2015imagenet}. Unless otherwise specified, the pre-training procedure elapses $300$ epochs with a base learning rate starting with $1.5\times10^{-4}$ and decays following the cosine annealing schedule. We only use normal data augmentations, including random cropping and horizontal flipping, to generate the $224\times224$ image (\textit{i.e.}, the target output). The image degradation includes:

\begin{itemize}
\item \textbf{(m) Input is a \texttt{masked} image.} This tasked is often referred to as masked image modeling (MIM). The setting follows MAE~\cite{mae}, where we set a ratio (\textit{e.g.}, $75\%$), randomly choose the portion of tokens, and remove them from the input image. The removed tokens are added back before $f^\mathrm{PT}\!\left(\cdot\right)$ with mask tokens and corresponding position embeddings.
\item \textbf{(a) Input is a \texttt{zoomed-in} image.} A part of the original image is zoomed in to occupy the entire canvas, and the remaining part is discarded. To make the task doable, the center coordinate and zooming ratio are fixed during the entire pre-training stage. We fix an integer $S$ ($S$ is a multiple of $16$ -- the token size), crop the central $S\times S$ region from the $224\times224$ image\footnote{Since the input cannot see the entire image, $S$ must be fixed, otherwise the learning task is unpredictable -- please check the difference from Task (b), where the input sees all necessary contents and thus $S$ and the location $(a,b)$ can be randomly sampled.}, and rescale it into $224\times224$ as input.  That said, the region outside the central $S\times S$ are invisible and the target model is required to recover it.
\item \textbf{(b) Input is a \texttt{zoomed-out} image.} The input image is down-sampled into a smaller scale so that it can be totally put inside the original canvas. The empty part is filled up with mirrored image contents so that the entire canvas look continuous (see Figure~\ref{fig:objectives}) . \textcolor{black}{We randomly sample an integer and a pair of coordinates $\left(a,b\right)$ to guarantee that a $S\times S$ square with the top-left corner at $\left(a,b\right)$ is totally located within the canvas}, rescaling the $224\times224$ original image into $S\times S$, and putting it at the corresponding location. 
%The outer region of the input image is filled up with mirrored image contents .
\item \textbf{(c) Input is a \texttt{distorted} image.} The original image is fed into a fisheye distortion algorithm which can cause a small portion of the image missing from input. 
% Given the center coordinate and degree of distortion, the recovery task is solved by simply mapping and interpolating image pixels. 
The degraded image is determined by the center coordinate and degree of distortion.
We adopt the \textcolor{black}{standard fisheye algorithm}. The center coordinate is uniformly sampled from the canvas, and the twist ratio is uniformly chosen between a small interval. The twist ratio impacts the pre-training performance (see Table~\ref{tab:diagnosis}).
\item \textbf{(d) Input is a \texttt{blurred} image.} The input image is blurred by applying a standard convolution (without bias) on each of the RGB channels individually. No further degradation (\textit{e.g.}, spatial transformations) is performed. The kernel size (square) is fixed but the convolutional weights are randomly sampled each time.
\item \textbf{(e) Input is a \texttt{de-colorized} image.} The input image is converted from RGB to grayscale, and no further degradation is performed. This task was verified successful for convolution-based networks~\cite{zhang2016colorful,larsson2016learning}, implying that image color is correlated to semantics. \textcolor{black}{In practice, we convert each image from RGB to HSV, multiply the S (saturation) channel by a value in $[0\%,30\%,60\%]$, and convert it back to RGB. Note that setting saturation to $0$ is close to performing RGB-to-gray, \textit{i.e.}, $\mathbf{x}_\mathrm{gray}=0.299\cdot\mathbf{x}_\mathrm{R}+0.587\cdot\mathbf{x}_\mathrm{G}+0.114\cdot\mathbf{x}_\mathrm{B}$.}
\end{itemize}

\begin{table}[!t]
\centering
\fontsize{9.5}{11.0}\selectfont
\setlength{\tabcolsep}{2.0mm}
\begin{tabular}{l|ccc|c|c}
\toprule
\multirow{2}{*}{\textbf{Task}} & \multicolumn{3}{c|}{\textbf{Factors}} & \textbf{Pre-training} & \textbf{ImageNet-1K} \\
{} & IM & ST & SC & \textbf{Epochs} & classification acc. \\ \midrule
\textit{None} & -- & -- & --  & -- & 80.7 \\ \midrule
MoCo-v3~\cite{moco-v3}$^\dagger$  & -- & -- & -- & 300 & 83.0 \textcolor{orange}{(+2.3)} \\
DINO~\cite{dino}$^\dagger$     & -- & -- & -- & 400 & 83.3 \textcolor{orange}{(+2.6)} \\ \midrule
BEiT~\cite{beit}  & \cmark  & \xmark & \xmark & 300 & 83.0 \textcolor{orange}{(+2.3)} \\
(m) \texttt{masked}~\cite{mae}  & \cmark  & \xmark & \xmark & 300 & 82.9 \textcolor{orange}{(+2.2)} \\ \midrule
(a) \texttt{zoomed-in}     & \cmark & \cmark & \xmark & 300 & 82.7 \textcolor{orange}{(+2.0)} \\
(b) \texttt{zoomed-out}    & \xmark & \cmark & \xmark & 300 & 82.2 \textcolor{orange}{(+1.5)} \\ 
(c) \texttt{distorted}$^\ddagger$   & \xmark  & \cmark & \cmark & 300 & 82.1 \textcolor{orange}{(+1.4)} \\
(d) \texttt{blurred}      & \xmark & \xmark & \cmark & 300 & 81.8 \textcolor{orange}{(+1.1)} \\ 
(e) \texttt{de-colorized}   & \xmark & \xmark & \cmark & 300 & 81.4 \textcolor{orange}{(+0.7)} \\ \midrule
(m)+(a) integrated & \cmark & \cmark & \xmark & 100 & 83.0 \textcolor{orange}{(+2.3)} \\
(m)+(a) integrated & \cmark & \cmark & \xmark & 300 & 83.2 \textcolor{orange}{(+2.5)} \\
\bottomrule
\end{tabular}
\caption{Image classification accuracy (\%) of different pre-trained models. All the backbones are ViT-Base/16. In the left-hand part, \textit{None} means no pre-training is used, and $^\dagger$ indicates image-based pre-training with contrastive learning. In the middle part, we list three factors of the learning objectives, including information missing (IM), spatial transformation (ST), and style change (SC), respectively. Please refer to the main text for further analysis. $^\ddagger$: the standard fisheye distortion incurs information missing (see Figure~\ref{fig:objectives}). To get rid of it, we distort a large image ($288\times288$) and set the prediction goal to be the central part ($224\times224$) -- this guarantees no information missing. Besides this table, we use the default setting which reports a $82.5\%$ accuracy (see Table~\ref{tab:diagnosis}).}
\label{tab:classification}
\end{table}

\subsection{Image Classification and Diagnosis}
\label{experiments:classification}

Following the convention~\cite{mae,beit}, we fine-tune the pre-trained models for image classification on ImageNet-1K (the same dataset used for pre-training). We adopt the regular setting for fine-tuning: the class-token features (with $768$ dimensions) produced by $f^\mathrm{B}\!\left(\cdot\right)$ are extracted and further fed into a linear $1000$-dimension classifier. The fine-tuning procedure elapses $100$ epochs. An AdamW optimizer~\cite{adamw} is used with a layer-wise learning rate decay and a weight decay of $0.05$.

Results are summarized in Table~\ref{tab:classification}. We find that all the designed learning objectives in this paper surpass the random baseline (indicating they help visual representation learning). That said, MIM is not the only way that works for token-based pre-training -- recovering other types of missing information also contributes. In addition, we extract three factors from each learning objective, \textit{i.e.}, whether it causes (i) information missing, (ii) spatial transformation, or (iii) image style change; the presence/absence of the factors with respect to the pre-training tasks are summarized in Table~\ref{tab:classification}. We also diagnose these learning objectives by adjusting some hyper-parameters -- results are summarized in Table~\ref{tab:diagnosis}. From the comparison, we summarize some important points as follows.

First, we discuss the impact of three factors on downstream recognition. When we compare Task (m), Task (b), and Task (d)/(e), we find that Factor (i) contributes most significantly to token-based pre-training, Factor (ii) is moderate, and Factor (iii) is the weakest. We conjecture that Factors (ii) and (iii) may have been weakened by the domain change caused by spatial transformation and/or image style change (these perturbed images will not appear in the downstream task); in comparison, Factor (i) does not suffer the weakness\footnote{Thanks to the property of vision transformers, each token is relatively individual. Even when the upstream pre-training is performed on partially masked images, the features extracted on each patch and the relationship between the visible patches will not change very much. Therefore, adding spatial masking, even at a high mask ratio (\textit{e.g.}, $75\%$), does not cause a significant domain gap between upstream pre-training and downstream fine-tuning.}, and thus the downstream recognition accuracy is relatively high.

In addition, we hope to stress that the three factors are often insufficient to determine the performance, and implementation details matter. For example, Task (d) is stronger than Task (e) although both of them have Factor (iii) only. Besides, even using Task (m), spatial masking, different sampling methods of masked patches can lead to varying performance -- the number ($82.9\%$) reported in Table~\ref{tab:classification} is obtained from the `random' setting (\textit{i.e.}, the presence of each patch is individually sampled), while the number is only $82.3\%$ when the `block' setting (\textit{i.e.}, a continuous region of patches are present while others masked) is used. That said, when improperly implemented, Factor (i) can be much weaker. Here, we discuss the strength of these factors based on the known best practice, while we keep an open eye to the future research that stronger implementations of Factors (ii) and (iii) may appear.

Next, we reveal the complementary nature of Factors (i) and (ii) by observing Tasks (a), which owns the `block' setting of Factor (i) and the normal setting of Factor (ii). While individual factors produce relatively lower performance (\textit{i.e.}, $82.3\%$ and $82.2\%$), combining them into Task (a) reports $82.7\%$, indicating that Factors (i) and (ii) can cooperate towards better recognition performance.

Motivated by the complementariness, we combine the best practice of these two factors towards stronger pre-trained models. 
% The implementation is straightforward: either degradation method (zooming-in and spatial masking) is considered an operator, and
We first perform zooming-in, then perform spatial masking -- note that swapping the order (masking then zooming-in) will result in the masked regions not occupying entire input tokens. As shown in Table~\ref{tab:classification}, the model pre-trained with the integrated learning objective outperforms ones pre-trained with individual tasks. Considering current results are quite high, $0.3\%$ accuracy promotion is notable. In addition, by pre-training for only 100 epochs, the integrated version matches MAE (300e) (\textit{i.e.} 83.0\% vs 82.9\%), which indicates its learning efficiency.
The good performance brought by integrating complementary learning objectives generalizes in two aspects. First, when the ViT-large model is used, the integrated task achieves a $84.7\%$ accuracy, surpassing that using the individual tasks by $0.3\%$. Second, as we shall see in Section~\ref{experiments:detseg}, the integrated model claims larger advantages in more challenging downstream tasks, that is, detection and segmentation.

\begin{table}[!t]
\centering
\fontsize{9.5}{11.0}\selectfont
\setlength{\tabcolsep}{2.5mm}
\begin{tabular}{l|c|ccccc}
\toprule
\textbf{Task} & \textbf{Setting} & \multicolumn{5}{c}{\textbf{Parameter \& Result}} \\ \midrule
%\multirow{2}{*}{Baselines} & \textit{name} & None & MoCo-v3 & DINO & BEIT & MAE \\
%{} & \textbf{IN-1K} acc. & 80.7 & 83.0 & 83.3 & 83.0 & 82.9 \\ \midrule
\multirow{2}{*}{(a) \texttt{zoomed-in}} & \textit{scale, $S$} & 96 & 128 & 160 & 192 & 208 \\
{} & \textbf{IN-1K} acc. & 82.2 & 82.3 & \textbf{82.7} & 82.5 & 82.3 \\ \midrule
\multirow{2}{*}{(b) \texttt{zoomed-out}} & \textit{scale, $S$} & 96 & 128 & 160 & 192 & rand. \\
{} & \textbf{IN-1K} acc. & 81.9 & 82.0 & \textbf{82.2} & 82.1 & 82.1 \\ \midrule
\multirow{2}{*}{(c) \texttt{distorted}} & \textit{twist ratio}  & \multicolumn{2}{c}{$.15$--$.20$} & $.20$--$.25$ & \multicolumn{2}{c}{$.25$--$.30$} \\
{} & \textbf{IN-1K} acc. & \multicolumn{2}{c}{\textcolor{black}{82.3}} & \textbf{\textcolor{black}{82.5}} & \multicolumn{2}{c}{\textcolor{black}{82.3}} \\ \midrule
\multirow{2}{*}{(d) \texttt{blurred}} & \textit{kernel size} & $5\times5$ & $7\times7$ & $9\times9$ & $11\times11$ & rand. \\
{} & \textbf{IN-1K} acc. & 80.9 & 81.5 & \textbf{81.8} & 81.0 & 81.7 \\ \midrule
\multirow{2}{*}{(e) \texttt{de-colorized}} & \textit{saturation} & \multicolumn{2}{c}{0\%} & 30\% & \multicolumn{2}{c}{60\%} \\
{} & \textbf{IN-1K} acc. & \multicolumn{2}{c}{81.2} & \textbf{81.4} & \multicolumn{2}{c}{81.3} \\
\bottomrule
\end{tabular}
\caption{The classification accuracy ($\%$) on ImageNet-1K (the pre-trained models are allowed to be fine-tuned), with different pre-training hyper-parameters. For the \texttt{zoomed-out} and \texttt{blurred} inputs, the random (rand.) setting indicates that each input image is pre-processed with a random scale or kernel size -- the value is uniformly sampled from the previous four options.}
\label{tab:diagnosis}
\end{table}

\subsection{Diagnostic Studies}
\label{experiments:diagnosis}

\begin{wrapfigure}{r}{0.5\textwidth}
\vspace{-2.5cm}
\centering
\includegraphics[width=6cm]{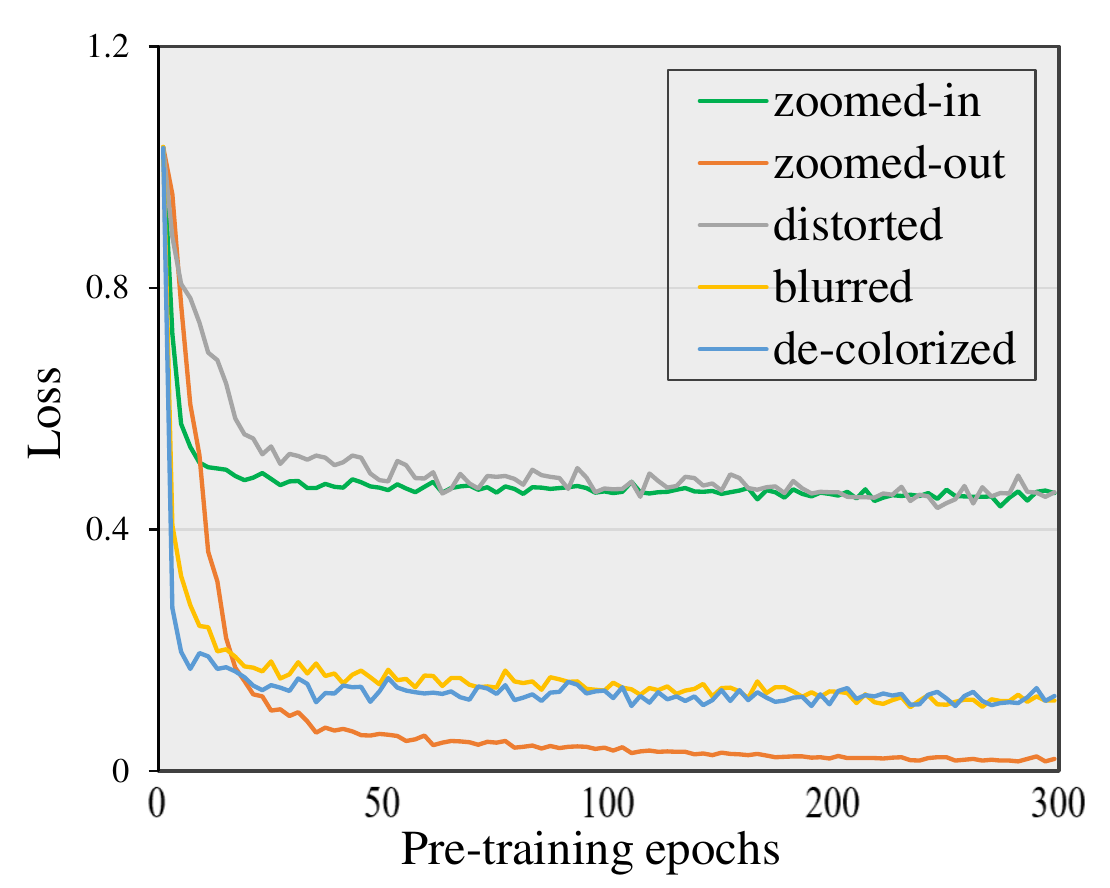}
\caption{The training loss curves of five learning objectives. The best hyper-parameter is used for each task.}
\vspace{-0.7cm}
\label{fig:curves}
\end{wrapfigure}

We first study how the difficulty of learning objectives impacts the pre-training performance. To be clear, the difficulty can be measured by the amount of missing information (\textit{e.g.}, the masking ratio) or the extent of spatial misalignment (\textit{e.g.}, the twist ratio). According to the diagnostic studies in Table~\ref{tab:diagnosis}, increasing the difficulty does not always bring benefits -- the best practice often appears with a moderate difficulty. This recalls the same conjecture that we have used in the previous subsection to compare the three factors: when the difficulty becomes arbitrarily large (\textit{e.g.}, the distortion is heavy), there is an inevitable domain gap between the images used in pre-training and fine-tuning, hence downgrading the performance of the pre-trained model. This opinion is further supported by the training loss curves of these learning objectives, shown in Figure~\ref{fig:curves}. Learning with Tasks (a) and (b) produces the best performance, while Task (a) reports the overall highest training loss and Task (b) the lowest. That said, difficulty is not the primary factor that impacts the pre-training performance.

Besides, we perform interesting studies to reveal other properties of token-based pre-training, \textit{e.g.}, the performance of Task (b) will be worse if we paste dummy patches to the outer region. Please refer to the supplementary material.

\begin{figure}[!t]
\centering
\includegraphics[height=10.5cm]{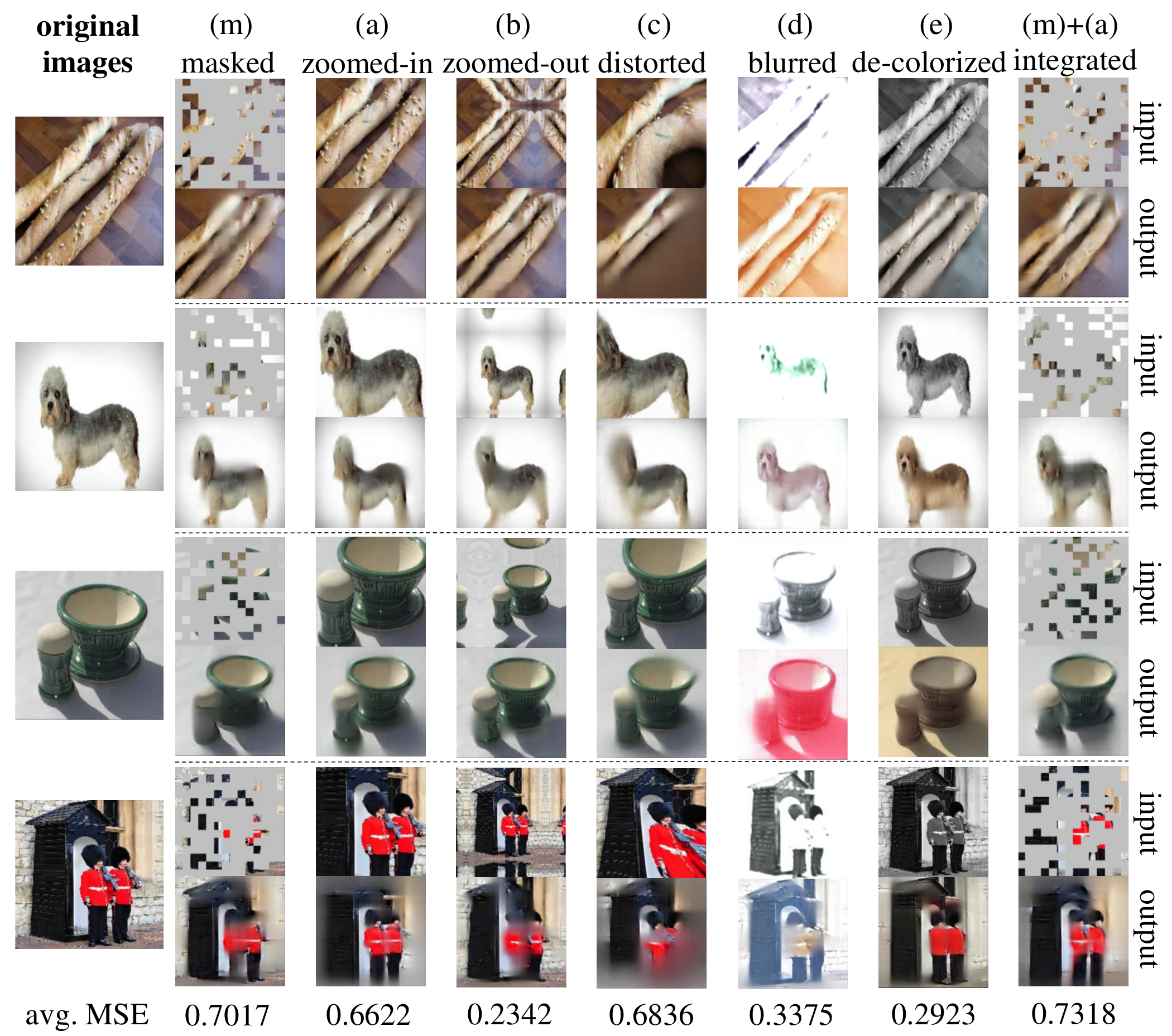}
\caption{Examples of recovered images on the ImageNet \textit{validation} dataset for MIM, the five proposed tasks, and an integrated task. The average MSE is computed on $100$ images. \textit{The figure is best viewed by zooming into details.}}
\label{fig:visualization}
\vspace{-0.2cm}
\end{figure}

Next, we visualize image recovery in Figure~\ref{fig:visualization} and offer the following analysis:
\begin{itemize}
\item All patches in the recovered images are more of less blurred, especially for the patches that are invisible to the input. In particular, for Task (b), although the \texttt{zoomed-out} input contains full image information, there is still significant blur on the output. This implies that the target model indeed encodes the image patches into visual features and then decode them for content recovery, not directly copying visual contents.
\item For Task (d), the input and output images may be largely different in color, but the average MSE value is much lower than Tasks (a) and (c) where the output images look more natural. We conjecture that the improper manipulation of color information can harm visual recognition, \textit{e.g.}, ImageNet contains a lot of fine-grained classes (\textit{e.g.}, birds) where color is crucial for classification. A possible way of alleviating the drawback is to equip the pixel-wise loss function with semantic guidance~\cite{peco,sage}.
\end{itemize}

\begin{table}[!t]
\centering
\fontsize{9.5}{11.0}\selectfont
\setlength{\tabcolsep}{2.0mm}
\begin{tabular}{l|c|cc|c}
\toprule
\multirow{2}{*}{\textbf{Task}} &
% \multirow{2}{*}{Epochs}
\multirow{2}{*}{\textbf{Token-Based?}} & \multicolumn{2}{c|}{\textbf{COCO}} & \textbf{ADE20K} \\
{} & {} & AP$^{box}$ & AP$^{mask}$ & mIoU \\ \midrule
\textit{Supervised} & -- & 46.9 & 41.5 & 47.0 \\ \midrule
MoCo-v3~\cite{moco-v3}$^\dagger$ & \xmark & 45.5 & 40.5 & 47.2 \\
DINO~\cite{dino}$^\dagger$     & \xmark & 46.8 & 41.5 & 47.2 \\ \midrule
BEiT~\cite{beit}  & \cmark & 39.5 & 35.9 & 45.5 \\
(m) \texttt{masked}~\cite{mae}  & \cmark & 45.3 & 40.2 & 45.6 \\ \midrule
(a) \texttt{zoomed-in}     & \cmark & 44.2 & 39.4 & 44.8  \\
(m)+(a) integrated & \cmark & 46.1 & 40.9 & 46.2  \\
\bottomrule
\end{tabular}
\caption{Object detection and instance/semantic segmentation results on COCO and ADE20K. All the numbers are in $\%$, the COCO metrics are by AP (box or mask), and the ADE metric is by mIOU. $^\dagger$ indicates multi-crop augmentation is used. Note that the integrated task requires the same computational overhead as MAE~\cite{mae}, which is about $4\times$ more efficient than MoCo-v3~\cite{moco-v3} and DINO~\cite{dino}.}.
\label{tab:detseg}
\end{table}

\subsection{Object Detection and Segmentation}
\label{experiments:detseg}

We also follow the convention to perform object detection and instance segmentation on COCO, and semantic segmentation on ADE20K. Unlike ImageNet-1K, these data are not seen during the pre-training stage. For the COCO experiments, we use the Mask R-CNN head~\cite{he2017mask} implemented by MMDetection~\cite{chen2019mmdetection}. The $1\times$ schedule ($12$ epochs) is used, with the learning rate starting with $3\times10^{-4}$ and decaying by $10\times$ after the $9$th and $11$th epochs, and the layer-wise decay rate being $0.75$. We apply multi-scale training (the short axis is between $480$ and $800$ pixels while the long axis does not exceed $1333$ pixels) and single-scale testing. For the ADE20K experiments, we follow BEIT~\cite{beit} to use the UperNet~\cite{upernet} head. There are $160\mathrm{K}$ iterations and the batch size is $16$. An AdamW optimizer is used and the learning rate is chosen from the better one of $3\times10^{-4}$ and $4\times10^{-4}$. The input resolution is $512\times512$, and no multi-scale testing is used.

Results are summarized in Table~\ref{tab:detseg}. Please refer to the supplementary details for detailed numbers (\textit{e.g.}, $\mathrm{AP}_{50}$, $\mathrm{AP}_\mathrm{S}$, \textit{etc.}). One can observe that, most often, different recognition tasks show a similar trend, \textit{i.e.}, a model that produces a higher ImageNet classification accuracy also reports a higher detection/segmentation accuracy on COCO and ADE20K. In addition, the accuracy gap among the models is enlarged (\textit{e.g.}, \textcolor{black}{(m) outperforms (a) by $0.2\%$ in ImageNet classification, but the advantage becomes $1.1\%$, $0.8\%$, $0.8\%$ for COCO detection/segmentation and ADE20K segmentation).} This is arguably due to the larger gap, domain-wise and task-wise, raises a bigger challenge for the pre-trained model.

Similarly as in image classification, we integrated two learning objectives, spatial masking and zooming-in, towards better downstream performance. Results are also shown in Table~\ref{tab:detseg}. Again, the integrated model surpasses the individual models consistently, and the advantages become larger, \textit{i.e.}, $0.8\%$, $0.7\%$, and $0.6\%$ for COCO object detection, COCO instance segmentation, and ADE20K semantic segmentation, respectively. This again validates the statement that downstream recognition with a larger domain/task gap can benefit more from a strong learning objective. 
%A similar trend also appears when the ViT-large backbone is used, which indicates the potential of the integrated learning objectives in applying to large-scale neural networks.

\section{Discussions: Token-Based vs. Image-Based Pre-Training}
\label{discussions}

\textcolor{black}{The previous parts investigated some interesting properties of token-based learning objectives and suggest an effective way of integrating multiple tasks. While token-based pre-training is a promising direction for vision transformers, in this section, we mainly discuss its weakness in learning strong semantic information. Throughout this section, we inherit the pre-trained models from Table~\ref{tab:classification}, \textit{i.e.}, the ViT-base backbone with $300$ or $400$ (only for DINO) epochs of pre-training.}

%From the above experiments, the benefit of token-based pre-training is made clear. However, in this section, we are interested in a discrepancy that emerges when the pre-trained models are evaluated in two different ways, namely, the fine-tuning protocol (as we used in the prior part) and the linear probing protocol (when the backbone weights are frozen after pre-training). Throughout this part, we use the ViT-base backbone and the pre-training procedure elapses $300$ epochs (except for DINO that uses $400$ epochs).

We start with revealing a discrepancy between token-based pre-training (\textit{e.g.}, BEIT~\cite{beit}, MAE~\cite{mae}, and our work) and image-based pre-training (in particular, contrastive learning, \textit{e.g.}, MoCo-v3~\cite{moco-v3} and DINO~\cite{dino}) methods. From Tables~\ref{tab:classification}, one can observe comparable performance on the ImageNet when the pre-trained backbone is allowed to be fine-tuned. However, in the linear probing test (\textit{i.e.}, the backbone is frozen), \textit{e.g.}, MoCo-v3~\cite{moco-v3} and DINO~\cite{dino} achieve high accuracy ($76.2\%$ and $77.3\%$), while BEIT~\cite{beit} and MAE~\cite{mae} report much lower numbers ($37.6\%$ and $61.5\%$). Our best model (\textit{i.e.}, integrating MIM with zooming-in) reports $63.1\%$, higher than BEIT and MAE, but still much lower than MoCo-v3 and DINO. That said, the token-based pre-training methods largely rely on fine-tuning the backbone weights to achieve high recognition accuracy.

From the perspective of applications, fine-tuning the pre-trained backbone is acceptable, but the unsatisfying performance of linear probing makes us rethink the effectiveness of token-based pre-training. We are wondering why the token-based pre-trained models are inferior in the linear probing test. There are at least two possible reasons. First, it is difficult for the token-based pre-trained models to extract semantic features; second, it is insufficient for a single linear layer to make use of the semantic features distributed in separate tokens. To discriminate between these possibilities, we design a new test named non-linear probing, which works by freezing the backbone weights (as linear probing does) and inserting a few transformer blocks between the backbone and the linear head. The non-linear probing has an advantages of detaching the contributions of upstream pre-training and downstream fine-tuning (like linear probing), yet it allows complex functions to be learned for specific purposes (like fine-tuning).

With $2$ inserted transformer blocks, the classification of MAE and the best model is improved from $61.5\%$ and
\textcolor{black}{$63.1\%$ to $63.3\%$ and $64.7\%$}, respectively, but there is still a significant deficit to the pre-trained models of MoCo-v3 and DINO. When the number of inserted transformer blocks is increased from $2$ to $4$, the accuracy is improved marginally ($<0.5\%$). These experiments support the first claim: token-based pre-training is relatively weak in capturing visual semantics. This indicates that the connection between image recovery and visual understanding is not necessarily strong. In the contrary, image-based pre-training (more specifically, contrastive learning) is required to distinguish if two views belong to the same image -- the task itself generates image-level representations that are more friendly to linear probing. In addition, the data augmentations to generate different views (\textit{e.g.}, random cropping) challenge the model's consistency in visual representation -- this is useful for image classification, but the token-based pre-training does not enjoy the same benefit.

The above analysis makes us conjecture that image-based and token-based pre-training approaches are complementary. It is not as superficial as that the former works better for linear probing while the latter works better for fine-tuning, but in the following important aspects.
\begin{enumerate}
\item Image-based pre-training extracts global image features (as if an image is a whole) while token-based pre-training mostly focuses on individual tokens (as if an image is composed of individual patches).
\item Image-based pre-training (contrastive learning) learns visual representation of an image by comparing it to other images, while token-based pre-training (recovering missing contents) learns visual representation by encoding the relationship between patches in each individual image.
\item Image-based pre-training (contrastive learning) suffers a conflict between view discrepancy and feature consistency -- that is, to challenge the target model, the algorithm must sample quite different views from a single image (otherwise, the pre-training task is too easy), but when the view discrepancy becomes too large, the assumption that they generate similar or at least highly related features may not hold. token-based pre-training does not have such a drawback.
\end{enumerate}
The advantages and disadvantages depend on the pre-training data and downstream tasks. For example, image-based pre-training fits ImageNet well because many images contain an iconic object and the background is clean, but it may encounter trouble in web images like COCO~\cite{lin2014coco} and WebVision~\cite{webvision}. Hence, we look forward to future research on integrating these two kinds of approaches, enabling image-based pre-training to deal with complex images, and injecting richer semantics into token-based pre-training.

\section{Conclusions}
\label{conclusions}

In this paper, we investigate the token-based pre-training of vision transformers, and point out that masked image modeling (MIM), though being popular currently, does not show significant advantages over other (seemingly simple) learning objectives. In particular, we show that spatial misalignment helps pre-training (while MIM does not consider it), and equipping MIM with spatial misalignment results in stronger pre-trained models. Nonetheless, most token-based pre-training approaches heavily rely on downstream fine-tuning, arguably because they have not learned rich semantics. We look forward to seeing more efforts on solving the essential problems we have discussed in Section~\ref{discussions}.

\bibliographystyle{splncs04}
\bibliography{egbib}

\appendix
\section{Appendix}
\subsection{Experimental Details}
In this part, we introduce more experimental details about pre-training of the proposed tasks. All the models use an encoder-decoder architecture, where the encoder has 12 blocks with a dimension of 768 and the decoder has 4 blocks with a dimension of 384. We adopt normalized target pixel in default, which boosts the performance as shown in ~\cite{mae}. All the results are attained by pre-training for $300$ epochs on $16$ V100 GPUs. Unless otherwise specified, we adopt the $75\%$ mask ratio for MIM pre-training. All the fine-tuning results share the same settings.

In default, the center part of the original image is cropped and then resized to $224\times224$ as input for the zoomed-in task. However, we later find that random cropping to a larger image (such as $320\times320$) and then picking the center $224\times224$ as input will bring a better result ($\sim 0.1\%$). This benefit may come from the gap narrowing between pre-training and fine-tuning. For zoomed-out, we first randomly resize the original image to $S \times S$, and then generate randomly $(a, b)$ (coordinate of the central location) to pad the image to $224\times224$ by mirror flipping. In the blurring task, we use a randomly initialized convolutional kernel to slide on the original image to generate a blurred image. It is worth noting that the kernel is initialized with a normal distribution -- we also try the kaiming normal~\cite{he2015delving} and xavier normal~\cite{glorot2010understanding}, but both of them achieve poor (even worse than no pre-training) performance.

\begin{figure}[!t]
\centering
\includegraphics[height=4.8cm]{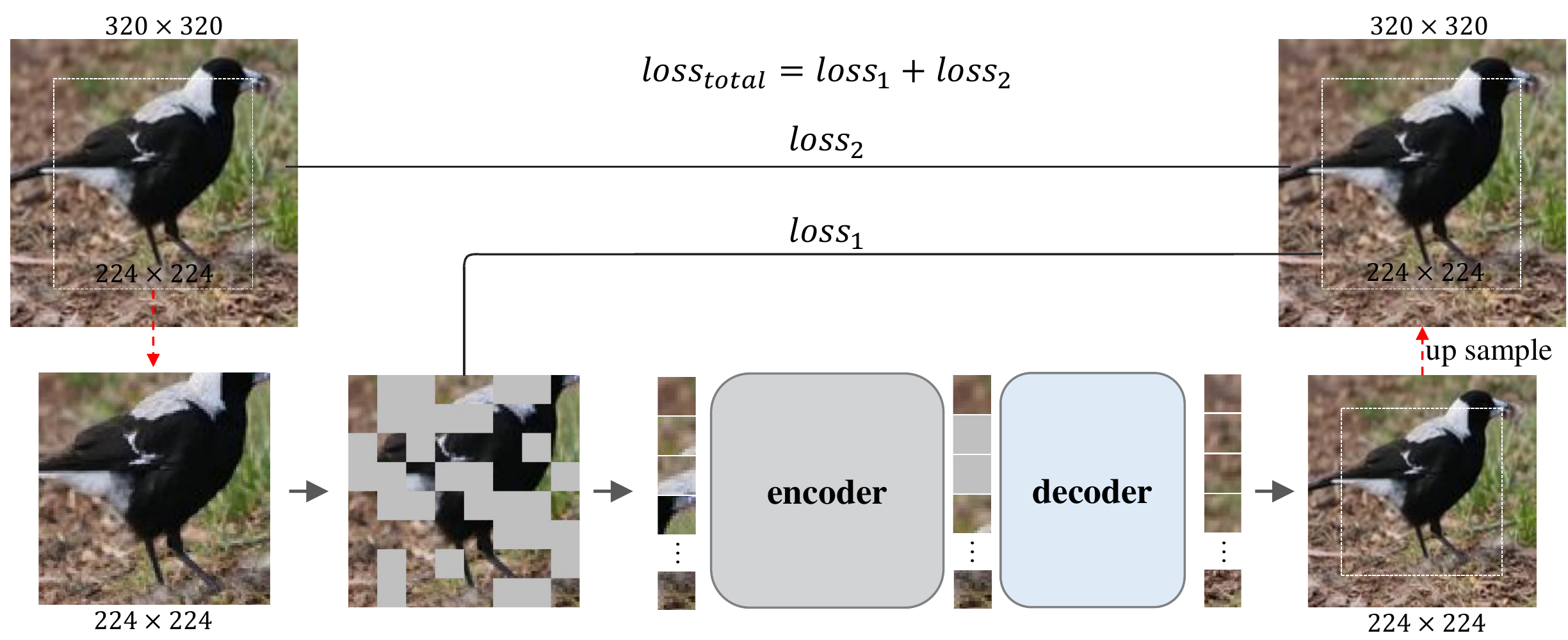}
\caption{Illustration of the details of integrating MIM and zoomed-in tasks.}
\vspace{-10pt}
\label{fig:integrate}
\end{figure}

\subsection{Details about the Integrated Task}
Fig.~\ref{fig:integrate} illustrates the implementation details of our integrated learning objective using MIM and zoomed-in. The raw image is first randomly cropped into $320 \times 320$, where the center $224 \times 224$ is picked as input. Before being fed into the neural network, the input is masked with a ratio of $75\%$, and the remaining patches are put into the encoder. A set of mask tokens are used to pad the masked patches corresponding to positions as shown in the pipeline. After padding, 196 tokens are sent into the decoder, which predicts the original image with a size of $224\times224$. The predicted image is then resized into $320\times320$ using the interpolation function with the mode of `area'. We supervise the predicted image using $2$ loss terms -- the predicted center part ($224\times224$) is supervised by the masked input, which is the same as ~\cite{mae}, and the outer band (except the center $224\times224$ part) is supervised by the outer band of the original $320\times320$ image. $2$ loss terms are jointly optimized during the pre-training process.

\subsection{More Diagnosis Experiments}

\textcolor{black}{We design 2-3 ablation experiments for task (a), (b) and (c) except (d) and (e) because of their poor performance. All the models are trained for $300$ epochs, and the results denote fine-tuning accuracies.}

\textcolor{black}{We propose $2$ settings to evaluate task (a). First, we randomly zoom-in a $160\times160$ part from the image instead of the center. The performance drops from $82.5\%$ to $81.5\%$ dramatically. Randomly zooming-in a local part as input makes the learning unstable, as how much the border region is taken from the original image is confusing for the transformer to predict.
% This also indicates that predicting the missing pixels is not a sufficient condition to train a self-supervised model.
Second, we take the inter $160\times160$ block as input and predict the outer patches using the MAE pipeline. The setting causes a result of $82.3\%$, which implies spatial misalignment is helpful to train self-supervised vision transformers.}

\textcolor{black}{For task (b), we add $3$ interesting experiments. We first only zoom out the original image to $160\times160$ without padding. This gets a poor performance of $79.9\%$. Then, we randomly pad the edges using black (empty) patches or patches from other images. These settings lead to $80.0\%$ and $80.5\%$ accuracies respectively. All these $3$ settings make the recovering task too easy to learn useful information. The first one is equivalent to just learn a formulation of image interpolation. The latter two settings provide distinct and sharp borders to the image, which is easy for the model to distinguish. 
% only require the model to learn the ability of magnifying borders between real patchs and noise patchs, which is irrespective of semantics.
}

\textcolor{black}{For task (c), we design $2$ settings. First, to guarantee there are no missing pixels for the input, we distort a $288\times288$ image but only predict the center $224\times224$ part. The distorted image is resized into $224\times224$ before being fed into the transformer. The result is $82.1\%$, just slightly lower than $82.5\%$, which illustrates that the improvement of task (c) comes from the distortion operation but not pixel missing. Second, we use an easier distorted transformation - horizontal wave distortion, which does not lead to pixel missing. Interestingly, we also achieve a good result of $81.9\%$, which further verified that recovering distorted images is able to improve self-supervised transformer.}

\begin{figure}[t!]
\centering
\includegraphics[height=4.8cm]{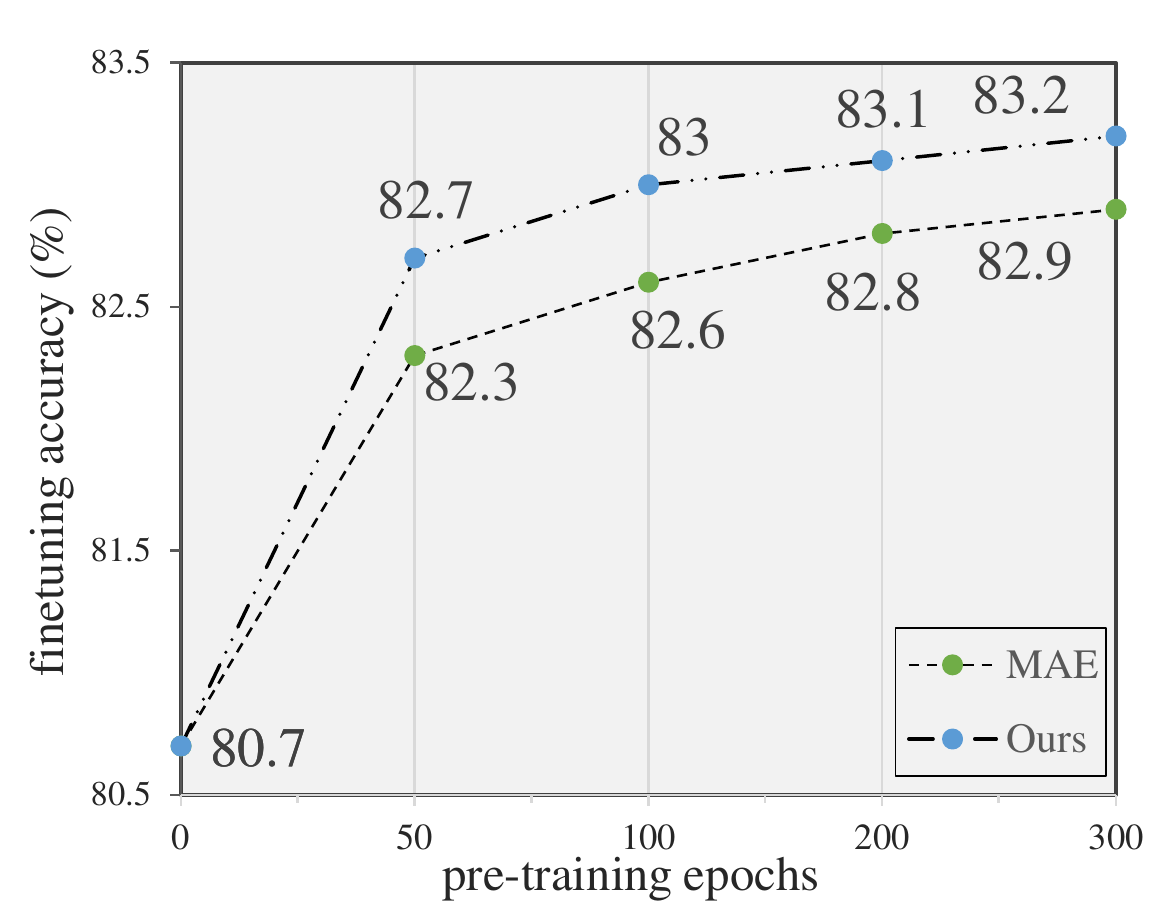}
\caption{Illustration of efficiency of the integrated task in this paper.}
\vspace{-10pt}
\label{fig:efficiency}
\end{figure}

\subsection{Analysis on the Integrated Task}
We first note that the integrated task combining MIM and zoomed-in does \textbf{not} require extra computational cost compared to MAE~\cite{mae}. Besides, the proposed zoomed-in task enjoys two issues. First, the zoomed-in task accepts the full image as input, which does not introduce a gap between pre-training and fine-tuning. Second, the zoomed-in task involves spatial misalignment, which brings greater challenges and helps the integrated task to be better. We pre-trained the integrated task for different epochs independently, including 50, 100, 200, and 300, and fine-tuning accuracies are shown in Fig.~\ref{fig:efficiency}. With pre-training only for $50$ epochs, our method achieves a competitive result of $82.7\%$. The score reaches $83.0\%$ by pre-training for only $100$ epochs, which matches MAE with pre-training for $300$ epochs. This demonstrates the high learning efficiency of the integrated task.

\end{document}